\documentclass[10pt,conference,onecolumn]{IEEEtran}

\usepackage{cite}
\usepackage{amsmath}
\usepackage{amssymb}
\usepackage{graphicx}
\usepackage{subcaption}
\usepackage{listings}

\title{GymFG: A Framework with a Gym Interface for FlightGear}
\author{
    \IEEEauthorblockN{Andrew Wood\IEEEauthorrefmark{1}, Ali Sydney\IEEEauthorrefmark{2}, Peter Chin\IEEEauthorrefmark{1}\IEEEauthorrefmark{2}, Bishal Thapa\IEEEauthorrefmark{2}, Ryan Ross\IEEEauthorrefmark{2}}
    \IEEEauthorblockA{\IEEEauthorrefmark{1}Department of Computer Science\\
                                           Boston University\\
                                           Boston, Massachusetts\\
                                           \{aewood, spchin\}@bu.edu}
    \IEEEauthorblockA{\IEEEauthorrefmark{2}Raytheon BBN\\
                                           Boston, Massachusetts\\
                                           \{asydney, pchin, bthapa, rross\}@raytheon.com}
}

\begin{document}
\maketitle

\begin{abstract}

    Over the past decades, progress in deployable autonomous flight systems has slowly stagnated. This is reflected in today's production air-crafts, where pilots only enable simple physics-based systems such as autopilot for takeoff, landing, navigation, and terrain/traffic avoidance. Evidently, autonomy has not gained the trust of the community where higher problem complexity and cognitive workload are required. To address trust, we must revisit the process for developing autonomous capabilities: modeling and simulation. Given the prohibitive costs for live tests, we need to prototype and evaluate autonomous aerial agents in a high fidelity flight simulator with autonomous learning capabilities applicable to flight systems: such a open-source development platform is not available. As a result, we have developed GymFG: GymFG couples and extends a high fidelity, open-source flight simulator and a robust agent learning framework to facilitate learning of more complex tasks. Furthermore, we have demonstrated the use of GymFG to train an autonomous aerial agent using Imitation Learning.  With GymFG, we can now deploy innovative ideas to address complex problems and build the trust necessary to move prototypes to the real-world. 
\end{abstract}

\section{Introduction}



Despite considerable advances in autonomous aerial research and the existence of advanced autonomous features in more capable air-platforms, only simple physics-based systems such as auto-pilot-based navigation are deployed/enabled in production. One critical contributing factor is the lack of trust in such systems. To address trust, we must first revisit the process for developing autonomous capabilities. 

Today, modeling and simulation is the premier process for research and development of autonomous capabilities both in industry and academia. There are several high fidelity flight simulators, and optimization and learning frameworks~\cite{SDLA2017,prepar3d2019,TRSP2014,DSDP2019}. However, we lack an open-source, autonomous-learning aerial development platform that combines a high fidelity flight simulator, a learning toolkit, and a robust optimization framework that supports the class of actions found in flight operations. To this end, we propose GymFG. GymFG adapts and extends the high fidelity FlightGear simulator and OpenAI's Gym learning toolkit: two of the most popular and accepted open-source tools in both academia and industry \cite{flightgear, gym}. Additionally, GymFG integrates seamlessly with optimization frameworks such as Tensorflow~\cite{tensorflow2015-whitepaper}. This innovative autonomous-learning aerial platform bridges the gap between current simple physics-based agents to more complex autonomous agents.


\section{Background}
Reinforcement learning (RL) assumes that an agent exists in an environment. In particular, the agent takes an \textit{action} and observes a \textit{state} and \textit{reward} from the environment. RL algorithms seek to maximize the total reward observed by the agent by predicting the most rewarded sequence of actions. Over the past decade, RL research has shown that not only can agents solve difficult problems without task-specific engineering~\cite{mnih2015human,schulman2015trust,mnih2016asynchronous}, but that it can compete with humans in a variety of tasks~\cite{ACT2019,OpenAI_dota}, even those with long event horizons~\cite{dsilea,OpenAI_dota} (i.e. tasks where there are long stretches of time between events where there is minimal/undefined reward). In 2016, OpenAI-Gym was developed to expedite the process of RL research by standardizing how agents interact with their environment, and by changing the focus of RL research from competition to peer-review by hosting leaderboards for various tasks.

GymFG focuses on providing a platform for aerial RL research that uses an OpenAI-Gym interface to high-fidelity 3D flight simulation. GymFG provides a framework to launch simulations containing arbitrary aerial situations with aircraft providing arbitrary roles, all with an easy to use, plug and play design.

\section{Design Decisions}
The design of GymFG is based on making the integration between FlightGear, gym, and users as intuitive as possible. Below, we summarize several of our main design decisions:

\paragraph{Teams of aircraft} A core concept in GymFG is that each aircraft is assigned to a team, and a single episode consists of at least one team interacting with a shared environment. This design was adopted to allow users to configure aircraft and aircraft relationships during runtime and then apply that configuration to an environment. All teams in a simulation must have a unique team name, and all aircraft within a team must have a unique callsign. In addition to reducing computational complexity, this scheme has the benefit of aligning with real life flight organization.

\paragraph{Aircrafts observe restricted information about their opponents} The environment defines a set of state variables that comprise what information an aircraft views of their opponents. This design decision more realistically reflects real world operations where opponents do not willingly share information besides what is physically observable. As a result, the default state variables available are the position and orientation of all opponents.

\paragraph{Environments define workload distribution and visualization} There are multiple types of environments that define how the workload is distributed across machines and the number of teams that can interact:
\begin{enumerate}
    \item Offline environments: Simulations occur on the local machine and have no mechanism for remote teams to connect, nor for other local teams to connect after a simulation has started.
    \item Online $k$-fixed environments: Simulations are distributed across potentially multiple machines, but only $k$ teams are allowed to participate in a simulation while more teams are denied access. A simulation starts once all $k$ teams connect, and ends once a single team completes their tasks.
    \item Online open environments: Simulations are distributed across potentially multiple machines and allow an unbounded number of teams to connect. 
\end{enumerate}
In addition, the environment defines whether or not a simulation can be visualized. This is ideal for running large scale simulations for training, and then visualizing small test instances, as a visualization-enabled simulation opens network ports per aircraft which can overload local networks.

\paragraph{Centralized environment state} Each team reports the state of all member aircrafts to a centralized state server at each step (except for the offline environment, in which the centralized state is built directly into the environment). The centralized server is responsible for assembling each team's state into a central state, and then broadcasting this central state to all clients (representing a single team). The centralized state server defines the set of state variables used to construct opponent states, but to reduce computational load, each client is responsible to further process the central state to restrict opponent information. This design decision minimizes the amount of network connections necessary for distributed teams to communicate with each other, and creates minimal configuration for users.

\paragraph{Task classes define aircrafts' state, reward, and stopping criteria} Supplied with each aircraft per team is a task. By allowing a task to define a corresponding aircraft's state information, it can encapsulate not only the logic for extracting the state of the aircraft, but also encapsulate the logic for computing the reward and stopping criteria for that aircraft. Creating new tasks is as simple as defining a reward function, identifying terminal conditions, and deciding what information each aircraft can observe.

\paragraph{JSBSim aircraft simulation with FlightGear visualization} JSBSim~\cite{berndt2004jsbsim} is a high-fidelity flight dynamics model that FlightGear uses internally. By using JSBSim in the background to process individual aircraft simulations, GymFG can avoid overhead (including visualization) required to launch a FlightGear instance per aircraft. Additionally, if visualization is desired, FlightGear instances can be created and bound to use the output of JSBSim instances for simulation rendering. Finally, by decoupling FlightGear from JSBSim, multiple FlightGear visualization instances can be created per aircraft during runtime to display different views of the same aircraft.

\section{Features}
GymFG offers a large set of features designed for ease of use while maximizing the flexibility of the package for RL and Imitation Learning (IL) agents. Below is a list of features currently supported:

\paragraph{Aircraft relationships} Aircraft and associated reward functions are related through the Team interface. Specifically, a team is created by passing it a team name, a list of aircrafts and associated tasks for each aircraft, and finally a map specifying any additional aircraft used to compute each aircraft's reward signal. For example, here are some sample team configurations (written using python3.6 syntax):

\begin{lstlisting}

from gymfg.aircraft import Aircraft, f15
from gymfg.tasks import DummyTask
from gymfg import Team

# Create a team with one aircraft that does not use any other
# aircraft states when computing the reward signal
red_team: Team = Team(name="red", roster=[(f15("red-one"), DummyTask)],
                      reward_function_targets={f15("red-one"): []})

# Create a team with two aircraft where one aircraft uses
# the state of the 2nd aircraft when computing its reward
# and the other aircraft does not use another state for its reward
red_team: Team = Team(name="red", roster=[(f15("red-one"), DummyTask),
                                         (f15("red-two"), DummyTask)],
                      reward_function_targets={f15("red-one"): [f15("red-two")],
                                               f15("red-two"): []})

# Create multiple teams with one aircraft each, and where the aircraft
# in each team uses the states of all other aircraft to compute its reward
team_names: List[str] = ["red", "blue"]
all_aircraft: List[Aircraft] = [f15("%s-one") % name for name in team_names]
teams: List[Team] = [Team(name=name, roster=[(my_aircraft, DummyTask)],
                          reward_function_targets={
                            my_aircraft: [trg_aircraft for trg_aircraft in all_aircraft
                                                       if target_aircraft != my_aircraft]
                          })
                     for my_aircraft in all_aircraft]
\end{lstlisting}

By default, each reward function is passed the state of its corresponding aircraft as a argument, but additional aircraft states must be specified via the team config if needed. This feature allows users to specify arbitrary relationships between aircrafts in the simulation.

\paragraph{Episode time definition} The Team interface optionally allows users to specify how long, in seconds, an episode for their team should last. This number defaults to 90s, and ultimately controls the number of steps in an episode. The smallest specified episode time throttles the entire simulation for all teams involved, as support does not currently exist for teams that drop out and rejoin during runtime.

\paragraph{Gym interface} GymFG adopts the gym interface with a slight modification to the control loop. A typical GymFG control loop looks like (using python3.6+ typing syntax):

\begin{lstlisting}

team_state, opponent_states: Tuple[List[np.ndarray], np.ndarray] = env.reset(team)
done: bool = False

while not done:
    team_actions: List[np.ndarray] = list()

    for aircraft_state, agent in zip(team_state, AGENTS):
        team_actions.append(agent.act(aircraft_state, opponent_states))
    team_state, opponent_states, rewards, done, info = env.step(np.hstack(team_actions))

\end{lstlisting}

The slight difference between GymFG and gym is that after creating an environment, GymFG requires users to supply their custom team configuration to the environment as an argument to the \texttt{reset} method (which gym takes no arguments in gym). This argument is ignored if a team has already been configured via a previous call to \texttt{reset}, but otherwise is used to initialize the local simulation and, if necessary, connect to the centralized state server.

\paragraph{Reward function, geometry, orientation, and state variable submodules} This functionality has been separated into different submodules to maximize encapsulation. Creating these submodules increases the flexibility with which they can be combined to represent new tasks and custom reward functions. By creating state extraction functionality between geometry, orientation, and state variables, users can assemble custom reward functions by writing a single lambda-function to compare different aspects of states.

\paragraph{Controller Library} Along with the environment, we developed a small library of primitive physics-based controllers designed to solve small tasks regarding basic flight. Our intention for doing so is twofold: we hope that users creating IL and RL models might find it useful to have controllers that provide accurate flight controls with which to test their models against, and also to test GymFG. We provide six main controllers: reaching a target altitude (dynamically or statically determined), one controller for manipulating each control (aileron, elevator, rudder, and throttle) until a desired configuration is reached, and keeping a level flight path. Each controller is accompanied with several tasks: a static task where the target is fixed (e.g. ReachStaticTargetRollTask) and a dynamic task (e.g. ReachDynamicTargetRollTask) designed to incrementally allow autopilot functionality. These tasks were additionally used to test GymFG, and to demonstrate that controllers can maximize a reward function in GymFG. Below are some resulting metrics from executing two of the controllers mentioned above. Figure 1, on the left the execution of an elevator controller using the ReachStaticTargetAltitudeTask task. Each row represents a different episode with a different desired altitude, and each column shows the control signal, the target signal, and the reward signal from each episode. Note that the elevator controller does maneuver the aircraft to reach the desired altitude, and that the aircraft maintains stable flight afterwards. On the right is the execution of a roll controller using the ReachStaticTargetRollTask. Like the previous experiment, each row represents a separate episode, and each column represents a signal from the experiment. Note that the roll controller can turn the aircraft until the desired roll is reached, and that the desired roll is maintained afterwards.


\begin{figure*}[!h]
    \centering
    \begin{subfigure}{0.5\textwidth}
        \centering
        \includegraphics[width=\linewidth]{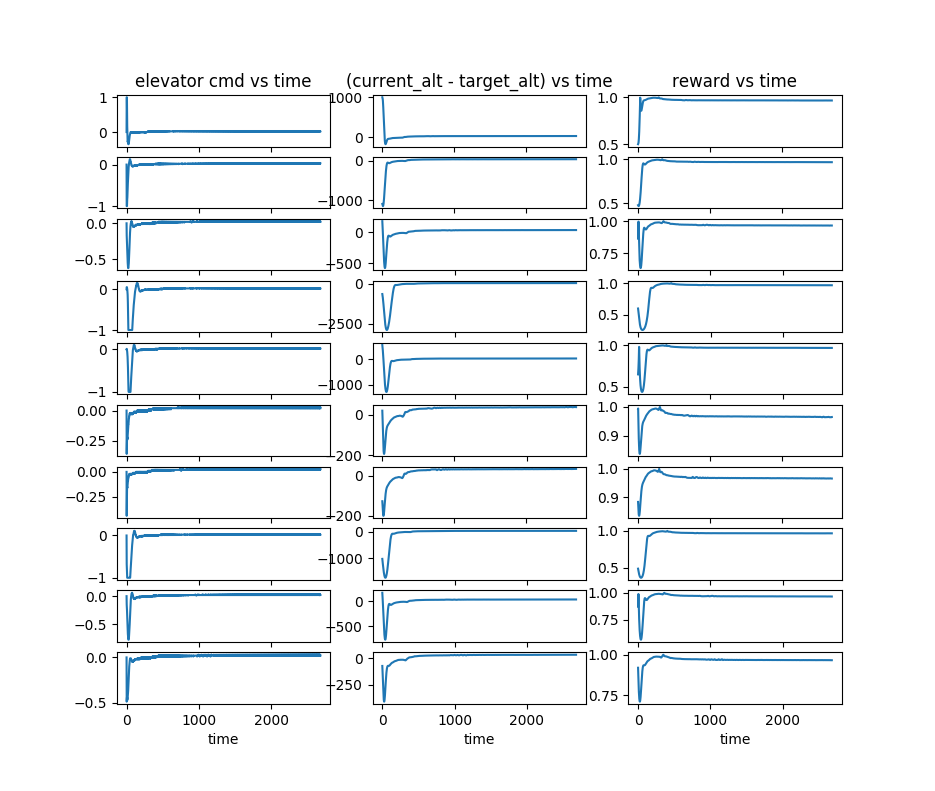}
    \end{subfigure}%
    \begin{subfigure}{0.5\textwidth}
        \centering
        \includegraphics[width=\linewidth]{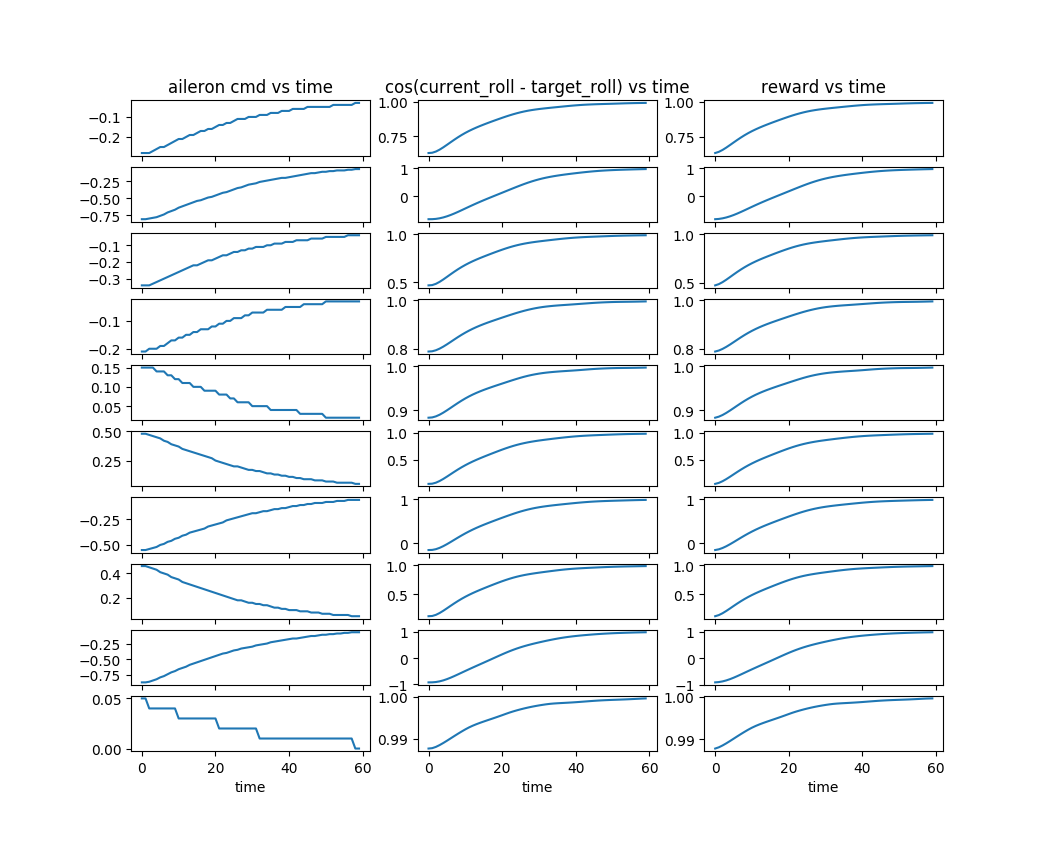}
    \end{subfigure}
    \caption{Two controllers: reach target altitude (left) and reach target roll (right). Note that rows in this figure represent different episodes.}
    \label{fig:controllers}
\end{figure*}

\paragraph{Reward function time sensitivity} Reward functions by default accept the previous state (after being processed by the environment to account for opponent views), the current state, the previous action of the aircraft, and the current action of the aircraft. This information can be disregarded or used as the user sees fit, allowing for reward functions to be a function of time as well as static reward functions. Future versions may support processing further timesteps of the environment's Markov decision process.

\paragraph{State and controls visualization} In addition to FlightGear visualization, GymFG also renders a small GUI that visualizes the control inputs as well as desired variables (i.e. typically all state variables). Control inputs are rendered graphically so users can intuitively see the position of the control stick as well as the throttle position, which display variables are aligned to show the name of the variable as well as the current value. This GUI is rendered every simulation step.

\paragraph{Manual flying} GymFG supports manual control. In certain circumstances, such as gathering data, it may be necessary to fly the aircraft manually. For example, consider an IL problem where the goal is for the aircraft to perform a complicated maneuver which is hard to control via a physics-based controller. To realize this objective, GymFG utilizes the ManualController object. ManualController signals from either a keyboard or a joystick as fed into the corresponding JSBSim instances.

\paragraph{Trajectory recordings} GymFG also provides the facilities to record (i.e. save to disk) sequences of states, rewards, and actions, known as a trajectory. Combining this functionality with manual flying, agents, and the physics-based controllers, GymFG provides the functionality to replay episodes, train IL models, and create custom visualizations of agent performance.

\newpage
\section{Future Directions}
There are several features that we intend to include with subsequent versions of GymFG. Most notably:

\paragraph{Online multiplayer} FlightGear is one of the most popular high-fidelity flight simulators in the world, and naturally supports multiplayer flight. FlightGear achieves this by hosting multiplayer servers which allows individual instances to connect and share data. We envision GymFG tapping into this functionality to allow agents to perform in environments against human players who are non the wiser about the artificial nature of their adversary.

\paragraph{Video capture} This feature is connected with enabling online multiplayer. GymfG's centralized state server provides the mechanism of serving state information between all teams in simulation, and does not have a way to connect with a FlightGear multiplayer server. Therefore, to collect state information on other aircraft in a multiplayer server, agents would have to process video frames from the simulation in real time. Future support will include the ability to receive video frames as input features without having to display the frames.

\paragraph{Arbitrary past reward functions} As mentioned previously, GymFG only allows reward functions to have access to the state and actions of the previous timestep. While solely using the past timestep is a typical feature of RL research, having the ability to examine past timesteps when computing problems with significant event horizons may be useful.

\paragraph{More aircraft options} Currently GymFG only supports three aircraft: f15(c), cessna172p, and a320. We envision including the available fleet to include all aircraft available to JSBSim and FlightGear, with the ability to create your own custom aircraft.

\paragraph{Joint initialization conditions} Each Task in GymFG currently provides its own initial conditions which are independent of all other aircraft in the simulation. Therefore, scenarios which require joint synchronization such as formation flying are difficult to simulate in GymFG. We envision the ability to override the initial conditions with a set of user-created initial conditions that would be significantly easier to jointly coordinate even under the presence of randomness.

\paragraph{Arbitrary team disconnect and rejoining} Allowing new teams to join ongoing simulations, allowing existing teams to disconnect during simulations, and allowing previously disconnected teams to reconnect would drastically improve the types of scenarios that GymFG could simulate. Such scenarios include, but are not limited to multi-party combat, reinforcements during combat, and development of fault tolerant systems such as intermittent connections to the game.

\paragraph{Mouse controlled ManualController} Manually flying with the keyboard can be difficult. A more intuitive way of flying would be to use the mouse as input to the elevator and ailerons. However, such control would still regulate the throttle and rudder controls to the keyboard.

\section{Acknowledgments}
This document does not contain technology or technical data controlled under either U.S. International Traffic in Arms Regulation or U.S. Export Administration Regulations. It has been approved for Public Release. Ref. No 20-S-0902, Approved 03/31/2020.

\bibliographystyle{IEEEtran}
\bibliography{main}

\end{document}